%% file: main.tex
\documentclass[10pt,twocolumn,letterpaper]{article}

\usepackage[pagenumbers]{wacv} 

\usepackage{graphicx}
\usepackage{amsmath}
\usepackage{amssymb}
\usepackage{booktabs}
\usepackage{pifont}     
\usepackage{bbding}     
\usepackage{fontawesome}
\usepackage{multirow}
\usepackage{colortbl}
\usepackage{subcaption}
\usepackage{mwe}
\usepackage[accsupp]{axessibility}

\usepackage{enumitem}

\usepackage{tcolorbox}
\usepackage{xcolor}   
\tcbuselibrary{theorems}
\newtcbtheorem[number within=section]{mytheo}{My Theorem}%
{colback=green!5,colframe=green!35!black,fonttitle=\bfseries}{th}
\definecolor{change}{rgb}{1,0,0}  
\definecolor{dark_gray_ours}{gray}{0.40}
\definecolor{light_gray_ours}{gray}{0.9}

\usepackage[pagebackref,breaklinks,colorlinks]{hyperref}

\usepackage[capitalize]{cleveref}
\crefname{section}{Sec.}{Secs.}
\Crefname{section}{Section}{Sections}
\Crefname{table}{Table}{Tables}
\crefname{table}{Tab.}{Tabs.}

\def\wacvPaperID{***} 
\def\confName{WACV}
\def\confYear{2025}

\definecolor{pink_ours}{rgb}{0.97, 0.64, 0.64}
\definecolor{blue_ours}{rgb}{0.62, 0.76, 0.90}
\definecolor{green_ours}{rgb}{0.53, 0.83, 0.64}
\definecolor{orange_ours}{rgb}{0.94, 0.68, 0.55}
\definecolor{purple_ours}{rgb}{0.82, 0.74, 0.96}
\definecolor{red_ours}{rgb}{1,0,0.09}
\definecolor{yellow_ours}{rgb}{1.0, 0.194, 0.78}
\definecolor{airforceblue}{rgb}{0.36, 0.54, 0.66}

\begin{document}

\title{ORID: Organ-Regional Information Driven Framework for \\ Radiology Report Generation}

\author{\textbf{Tiancheng Gu}$^\text{\ding{170}}$\footnotemark[1], 
\textbf{Kaicheng Yang}$^{\text{\ding{171}}}$\footnotemark[1] \\
\textbf{Xiang An}$^{\text{\ding{171}}}$, 
\textbf{Ziyong Feng}$^{\text{\ding{171}}}$, 
\textbf{Dongnan Liu}$^{\text{\ding{170}}}$, 
\textbf{Weidong Cai}$^{\text{\ding{170}}}$\footnotemark[2]\\
$^{\text{\ding{170}}}$University of Sydney $^{\text{\ding{171}}}$DeepGlint \\
\texttt\small{\small tigu8498@uni.sydney.edu.au, \small kaichengyang@deepglint.com}}

\maketitle

\renewcommand{\thefootnote}{\fnsymbol{footnote}}
\footnotetext[1]{Equal contribution.}
\footnotetext[2]{Corresponding author.}

\begin{abstract}
   The objective of Radiology Report Generation~(RRG) is to automatically generate coherent textual analyses of diseases based on radiological images, thereby alleviating the workload of radiologists. Current AI-based methods for RRG primarily focus on modifications to the encoder-decoder model architecture. To advance these approaches, this paper introduces an \textbf{O}rgan-\textbf{R}egional \textbf{I}nformation \textbf{D}riven~(\textbf{ORID}) framework which can effectively integrate multi-modal information and reduce the influence of noise from unrelated organs. Specifically, based on the LLaVA-Med, we first construct an RRG-related instruction dataset to improve organ-regional diagnosis description ability and get the LLaVA-Med-RRG. After that, we propose an organ-based cross-modal fusion module to effectively combine the information from the organ-regional diagnosis description and radiology image. To further reduce the influence of noise from unrelated organs on the radiology report generation, we introduce an organ importance coefficient analysis module, which leverages Graph Neural Network~(GNN) to examine the interconnections of the cross-modal information of each organ region. Extensive experiments and comparisons with state-of-the-art methods across various evaluation metrics demonstrate the superior performance of our proposed method.
\end{abstract}

\input{sections/introduction}

\input{sections/relatedworks}
\input{sections/method}
\input{sections/experiments_setting}
\input{sections/experiment_result_and_comparison}
\input{sections/conclusion}
\input{sections/appendix}

{\small
\bibliographystyle{ieee_fullname}
\bibliography{egbib}
}

\end{document}

%% file: sections/introduction.tex
\section{Introduction}
\label{sec:introduction}

\begin{figure}[ht]
\begin{center}
\includegraphics[width=1.0\linewidth]{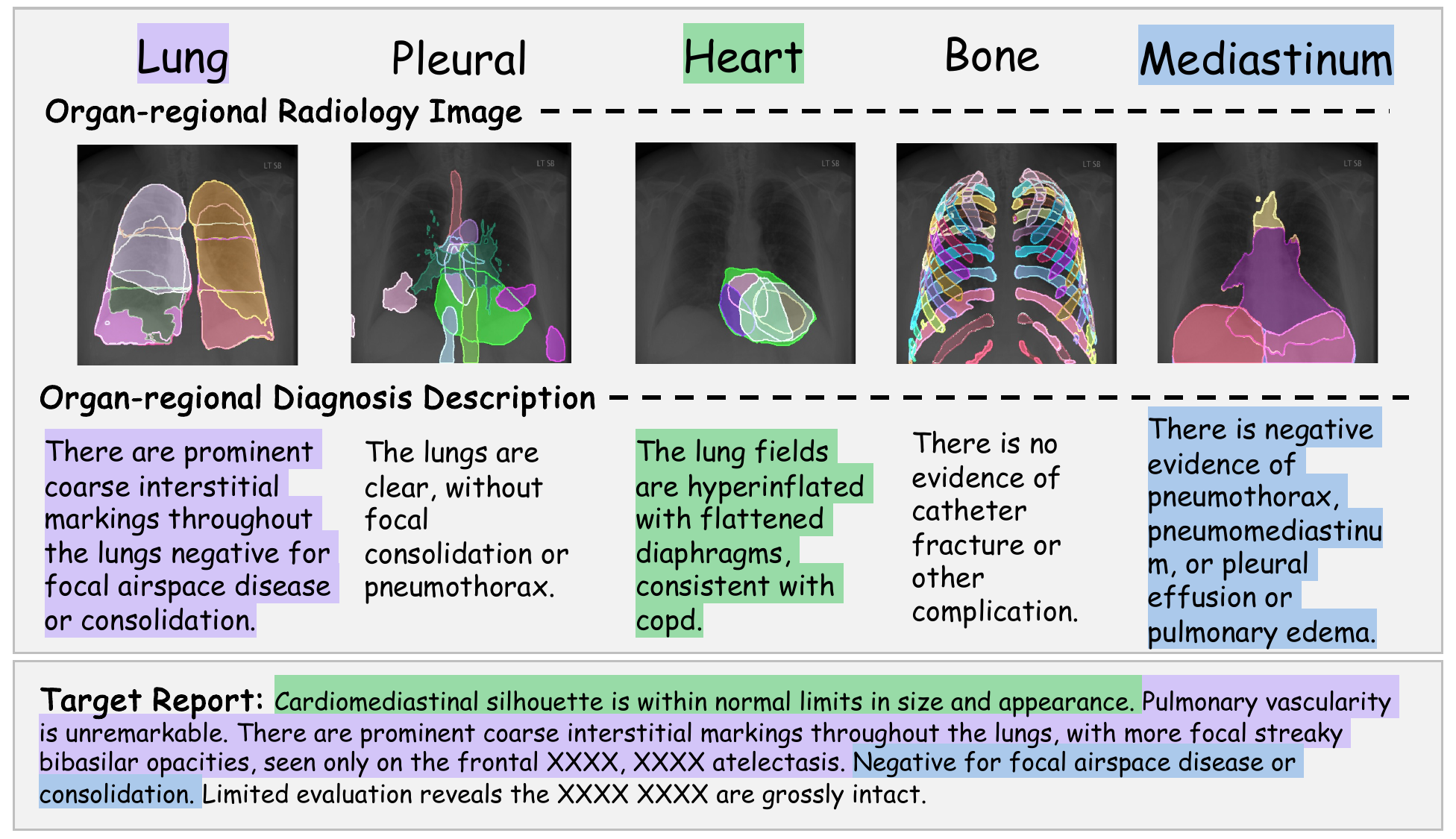}
\end{center}
\vspace{-0.5cm}
\caption{Visualization of organ-regional radiology image and diagnosis descriptions. Relevant segments associated with the target report have been highlighted using distinct colors.}
\label{fig:introduction}
\vspace{-0.5cm}
\end{figure}

Analyzing radiological images plays a crucial role in disease identification. The automatic generation of precise radiology reports can significantly alleviate the substantial workload faced by radiologists~\cite{rrg_important}. Manual analysis of radiological images is not only complex and expensive but also prone to errors, as radiologists need to examine numerous images and create detailed reports for each one~\cite{R2Gen}. In recent years, deep learning techniques have been extensively applied to generate textual descriptions for images, such as image captions~\cite{flamingo, meshed-memory, ALBEF}, inspiring many RRG methods~\cite{rgrg, DCL, kiut}. Nevertheless, radiology report generation differs from standard image captioning due to the intricate nature of radiological images, which typically highlight small, disease-specific regions \cite{iu_xray}. Additionally, radiological images primarily vary in the specific areas associated with diseases. This complexity extends to the textual descriptions in RRG, where the primary divergence lies in the analysis of disease findings, while the descriptions of healthy tissue remain relatively uniform~\cite{Xpronet}.

To address these challenges, MIMIC-CXR \cite{mimic_cxr} and IU-Xray \cite{iu_xray} have been introduced, which comprise nearly 400,000 and 7,000 image-report pairs. As a result, diverse RRG methodologies have been introduced~\cite{R2Gen, R2GenCMN, R2GenCMMRL, DCL, repsnet, jing-etal-2018-automatic, yang-etal-2021-writing, hou-etal-2023-recap, hou-etal-2023-organ, Gu_2024_WACV}, each tackling the issue from distinct angles and showcasing notable enhancements in performance. Existing methods can be broadly classified into two categories, the first category~\cite{R2Gen, R2GenCMN, R2GenCMMRL} concentrates on improving model architectures to effectively fuse image and text modalities for radiology report generation. The second category~\cite{DCL, repsnet, rgrg} utilizes external modality information and existing knowledge to bolster encoder-decoder models in producing conclusive reports. Despite their notable achievements, these methods do not incorporate detailed organ-regional information, which is crucial for guiding the generation of comprehensive radiology reports~\cite{importance_disease_detection, kumar2024improving}.

In this paper, building upon offline generated organ-regional radiology images and diagnosis descriptions~(as shown in Fig.~\ref{fig:introduction}), we propose an \textbf{O}rgan-\textbf{R}egional \textbf{I}nformation \textbf{D}riven~(\textbf{ORID}) framework for generating accurate and believable radiology reports. The ORID framework primarily comprises an organ-based cross-modal fusion module and an organ importance coefficient analysis module. Specifically, motivated by the exceptional performance of LLaVA-Med in medical diagnosis~\cite{llava-med}, we construct a radiology image based instruction dataset grounded in radiology images encompassing 10,000 question-answer pairs related to 4,000 radiological images. Then, this dataset is utilized to enhance organ-regional diagnosis description ability and facilitate the development of the LLaVA-Med-RRG model, which is employed for generating descriptions for downstream tasks.
After that, we propose an organ-based cross-modal fusion module to integrate the information from the organ-regional radiology image and diagnostic description. To reduce the influence of noise from unrelated organs, we utilize a prior-knowledge graph~\cite{kiut} to construct an adjacency matrix based on the relationships of each organ region. Finally, the fused cross-modal features are combined with the raw radiology image features and input into an encoder-decoder model to generate the final radiology report. The contributions of this paper can be summarized as follows:

\begin{itemize}[noitemsep,topsep=0pt]
    \item We construct an RRG-related instruction dataset to improve organ-regional diagnosis description ability and get the LLaVA-Med-RRG model which is used to generate descriptions for downstream tasks.
    \item We propose an Organ-Regional Information Driven~(ORID) framework for generating accurate and believable radiology reports. It consists of an organ-based multi-modal fusion module and an organ importance coefficient analysis module which can effectively integrate the multi-modal information and reduce the influence of noise from unrelated organs.
    \item We perform extensive experiments and the results prove that our proposed ORID framework achieves new state-of-the-art performance on two public radiology report generation benchmarks.
\end{itemize}

%% file: sections/relatedworks.tex
\section{Related Works}
\label{sec:relatedworks}

\subsection{Image Captioning}

Traditional image captioning techniques rely on inputting image-text pairs into an encoder-decoder model to generate text descriptions based on the input images. Early works~\cite{donahue2015long, gu2017empirical, vinyals2015show} primarily utilize the Long Short-Term Memory Network~(LSTM) for the language component and Convolutional Neural Network~(CNN) for the image component. Recently, catalyzed by the Vision Transformer~(ViT)~\cite{dosovitskiy2020image}, attention-based models~\cite{transformer} have gained prominence in the field of image captioning. Subsequent researches~\cite{anderson2018bottom, cornia2020meshed, huang2019attention} focus on incorporating object detectors to identify and extract relevant image regions to assist caption generation. Additionally, further studies~\cite{cornia2020meshed, nguyen2022grit, pan2020x} improve the architecture of the underlying model to enhance the interaction and alignment between image and text modalities. In contrast to general visual language tasks, radiology images present unique challenges due to their subtle and detailed differences. These nuances require specialized approaches to accurately capture and describe the intricate information in radiological imagery, such as the work~\cite{DCL} based on prior knowledge.

\definecolor{pink_fig2}{rgb}{0.96, 0.89, 0.82}
\definecolor{green_fig2}{rgb}{0.88, 0.93, 0.84}
\begin{figure*}[!ht]   
\begin{center}
\includegraphics[width=1\linewidth]{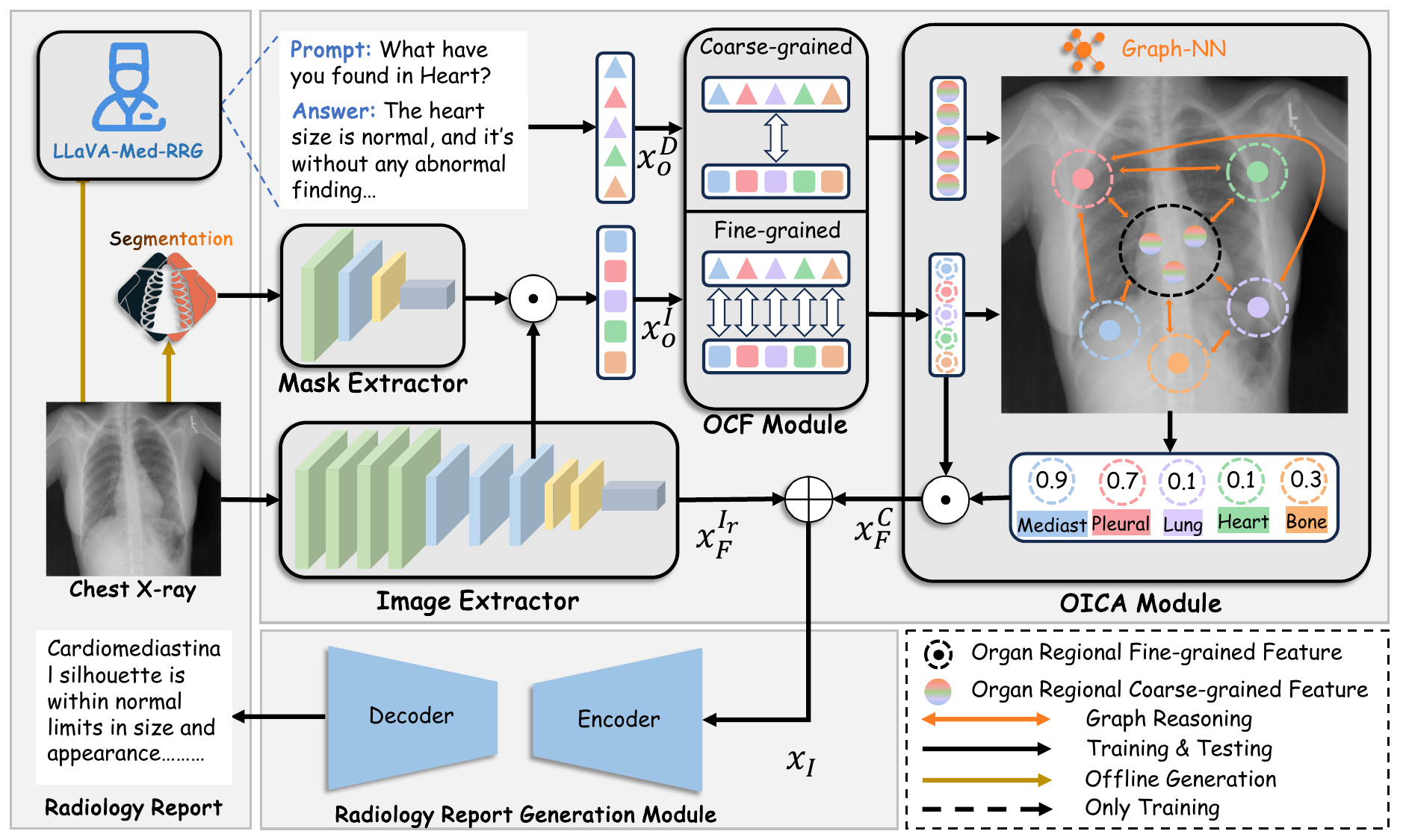}
\end{center}
\vspace{-0.5cm}
\caption{The overall architecture of our proposed ORID framework.}
\label{Fig: The structure of the model}
\vspace{-0.5cm}
\end{figure*}

\subsection{Radiology Report Generation}

Medical report generation extends the task of image captioning, presenting increased complexity and demanding high levels of accuracy and precision. Extensive works~\cite {R2Gen, R2GenCMMRL, DCL, repsnet, Gu_2024_CVPR, Gu_2024_WACV,  10230415, xu2024an} have led to significant advancements in this domain. Inspired by attention mechanisms~\cite{transformer}, the R2Gen~\cite{R2Gen} and R2GenCMN~\cite{R2GenCMN} enhance the encoder-decoder architecture by separately filtering and fusing image and text features using gate mechanisms from LSTM~\cite{LSTM} and cross-modal memory networks~\cite{R2GenCMN}. Building on these, the R2GenCMM-RL~\cite{R2GenCMMRL} further improves the R2GenCMN~\cite{R2GenCMN} by incorporating reinforcement learning-based reward mechanisms~\cite{RL}. In contrast to these models, the DCL~\cite{DCL} focus on structural improvements and it leverages a pre-generated knowledge graph with a correlation matrix of target report words. The RGRG~\cite{rgrg} employs bounding boxes to guide the model to assign more attention to abnormal regions which facilitates disease detection and report generation. Additionally, the RepsNet~\cite{repsnet} incorporates visual question answering knowledge~\cite{vqa_rad} to enhance the final report generation process. However, existing methods struggle to obtain precise organ-level analysis descriptions while underutilizing organ-regional information to improve the accuracy and relevance of generated radiology reports.

\subsection{Multimodal Large Language Model}

The advent of Large Language Models~(LLMs)~\cite{achiam2023gpt, touvron2023llama}, particularly those enhanced through instruction-tuning~\cite{longpre2023flan, mukherjee2023orca}, has demonstrated significant potential in serving as a versatile interface for language-related tasks. To advance the capabilities of LLMs beyond language perception, recent studies~\cite{xie2024croc, zhu2023minigpt, llava-med, liu2024visualllava} have expanded these models into Multi-modal Large Language Models~(MLLMs) by incorporating various modalities through instruction tuning~\cite{huang2024language, wu2024v}. Expanding upon these advancements, LLaVA-Med~\cite{llava-med} is formulated through the instruction-tuning of LLaVA~\cite{liu2024visualllava} using an extensive dataset of self-constructed, medically-oriented instructions. While LLaVA-Med exhibits strong performance in various medical domains, it lacks specific specialization in the field of radiology. To improve organ-regional diagnosis description ability, we construct an organ-level RRG-related instruction dataset that includes 10k question-answer pairs to develop the LLaVA-Med-RRG.

%% file: sections/method.tex
\section{Method}
\label{sec:method}

The overview architecture of our proposed \textbf{O}rgan-\textbf{R}egional \textbf{I}nformation \textbf{D}riven~(\textbf{ORID}) framework is shown in Fig.~\ref{Fig: The structure of the model}. This section introduces our primary methodologies, including the development of LLaVA-Med-RRG~(Sec.~\ref{subsection: LLaVA-Med-RRG}), the details of the Organ-based Cross-modal Fusion~(OCF) module~(Sec.~\ref{subsection: Global-Local Crosss-modal Feature Fusion Module}), the Organ Importance Coefficient Analysis~(OICA) module~(Sec.~\ref{subsection: Organ Importance Coefficient Analysis Module}), and the Radiology Report Generation Module~(Sec.~\ref{subsection: Radiology Report Generation Module}).

\subsection{LLaVA-Med-RRG}
\label{subsection: LLaVA-Med-RRG} 

Even though LLaVA-Med shows superior performance in analyzing general medical images~\cite{llava-med}, its performance with radiology images has been relatively less satisfactory. Motivated by the cost-effectiveness and high performance of instruction tuning~\cite{huang2024language}, we create an organ-level RRG-related instruction dataset to conduct instruction tuning on LLaVA-Med, resulting in LLaVA-Med-RRG. This enhancement is intended to enhance the capacity for organ-level analysis of radiology images. Specifically, the instruction dataset is transfered from IU-Xray~\cite{iu_xray} and MIMIC-CXR~\cite{mimic_cxr}. Drawing from the prior knowledge~\cite{kiut} of the Disease Symptom Graph~(DS-Graph)~(more details in supplementary materials), we identified that all diseases in radiology images are directly related to five organs: bone, pleural, lung, heart, and mediastinum. Building upon this prior knowledge, we use the NLTK tool\footnote{\url{https://github.com/nltk/nltk}} to segment captions into sentences and assign them to the corresponding organs by referring to the DS-Graph. The input and output type during the instruction tuning process is shown in Fig.~\ref{tab:prompt1}.
\input{tables/instruction_tuning_example}

To ensure the quality and diversity of the instruction dataset, we adhere to the following four rules during the data construction process: (1)~Enhancing positive disease analysis: increase the number of Question-Answer~(QA) pairs involving disease analysis while decreasing those without any disease~(standard organ description) to enhance the model's disease-analysis capabilities. (2)~Minimizing redundancy: reduce examples with redundant identical answers to promote diverse responses to similar questions. (3)~Prioritizing image examples with rich diagnosis: Concentrate on image examples containing more organ-specific QA pairs. (4)~Balancing organ analysis proficiency: maintain an equitable distribution of QA pairs across the five organs to prevent the model from exhibiting bias towards specific organ analyses. Following the construction of the dataset, we conduct standard instruction tuning on the pre-trained LLaVA-Med using LLaMA-Factory~\cite{zheng2024llamafactory}, resulting in the development of LLaVA-Med-RRG.

\subsection{Organ-based Cross-modal Fusion Module}
\label{subsection: Global-Local Crosss-modal Feature Fusion Module}

\paragraph{Data Flow.} After obtaining the LLaVA-Med-RRG model, we use it to generate organ-regional diagnosis descriptions and each radiology image corresponds to five description texts for distinct anatomical regions: lungs, heart, pleural, bones, and mediastinum. Then, we tokenize and embed these descriptions as features $x^{D}_{\text{o}}$, where \text{o} $\in \{\text{mediastinum}, \text{pleural}, \text{lung}, \text{heart}, \text{bone} \}$.

Radiology images often emphasize small, distinct regions within the image~\cite{Xpronet}. To address this characteristic, we utilize organ-regional information to assist the model in improving the analysis ability of radiology images. Specifically, we employ the CXAS~\cite{cxas} model to identify and segment specific regions of interest (such as the left lung, lung ribs, and scapula) within radiology images. Drawing upon prior knowledge that radiology reports often center around five specific organ regions~\cite{kiut}, we choose the pertinent segmented mask images and classify them into five distinct sets representing the heart, bones, lungs, pleura, and mediastinum, then we concat them in channel dimension in each set ~(more details in supplementary materials). After that, we integrate these masks with original image information to extract organ-regional image information $x^{I}_{\text{o}}$ as follows:
\begin{equation}
    \begin{aligned}
    & x^{I_{m}}_{\text{o}} = E_{m}(I_{m}), x^{I_{r}} = \hat{E_{r}}(I_{r}), \\
    & x^{I}_{\text{o}} = x^{I_{m}}_{\text{o}} \odot x^{I_{r}},
    \end{aligned}
\end{equation}
where $\odot$ is the element-wise multiplication. We use the mask feature extractor $E_{m}$(ResNet18) to extract the organ mask feature $x^{I_{m}}_{\text{o}}$ from the original radiology image $I_{m}$. Meanwhile, we feed the raw radiology image $I_{r}$ into the image feature extractor $E_{r}$(ResNet101) and extract the output of middle layer $\hat{E_{r}}$ as the raw radiology image feature.

\paragraph{Modality Fusion.} After acquiring the corresponding organ-regional image information $x^{I}_{\text{o}}$ and diagnosis description features $x^{D}_{\text{o}}$, we propose the organ-based cross-modal fusion module to fuse the image modality with the text modality to obtain a single fine-grained cross-modal feature $x^{C}_{o}$ in organ-level. The process is formally as follows:
\begin{equation}
\label{equ:MHA}
    \begin{aligned}
    &Q = x^{I}_{\text{o}}W^{Q}, K = x^{D}_{\text{o}}W^{K}, V = x^{D}_{\text{o}}W^{V}, \\
    &x^{C}_{\text{o}} = \text{MHA}(x^{I}_{\text{o}}, x^{D}_{\text{o}}) = \text{Softmax}(\frac{QK^T}{\sqrt{d_{n}}})V,
    \end{aligned}
\end{equation}
where MHA is the multi-head cross-attention~\cite{transformer} and $W^{Q}, W^{K}, W^{K}$ are learnable matrices.

Analyzing individual organ image features independently can capture the fine-grained organ-level difference, thereby improving organ disease analysis performance. At the same time, this approach may restrict the model's collaborative analysis capabilities, potentially leading to suboptimal analysis for diseases that rely on multiple organs~\cite{rrg_important}. To address this limitation, we generate a coarse-grained organ-level image feature $x^{I}_{T}$ by adding all organ image features, which can be indicated as follows:
\begin{equation}
    \small
    x^{I}_{T} = \sum_{i \in \phi} x_{i}^{I}, 
\end{equation}
where $\phi$ is the organ set \{mediastinum, pleural, lung, heart, bone\}. We also further generate corresponding coarse-grained organ diagnosis description features $x^{D}_{T}$ by using the following procedure:
\begin{equation}
    \small
    x^{D}_{T} = \text{Concat}(x^{D}_{\text{o}}) + \text{T}_{\text{p}} + \text{T}_{\text{o}},
\end{equation}
where $\text{Concat}(\cdot)$ is the concatenation operation, $\text{T}_{\text{p}}$ is positional embedding tokens, and $\text{T}_{\text{o}}$ is organ embedding tokens to distinguish different organs. Then, we integrate the coarse-grained organ image features~$x^{I}_{T}$ with its corresponding description features $x^{D}_{T}$ to get the organ-regional cross-modal information $x^{C}_{T}$ as follows:
\begin{equation}
    x^{C}_{T} =  \text{MHA}(x^{I}_{T}, x^{D}_{T}).
\end{equation}

\input{tables/SOTA_comparison_table}
\subsection{Organ Importance Coefficient Analysis Module}
\label{subsection: Organ Importance Coefficient Analysis Module}

Upon reviewing a substantial number of radiology reports, it is noted that most reports concentrate on analyzing 2-4 specific organ regions, with certain organs being reported as normal~\cite{iu_xray, mimic_cxr}. We hypothesize that including all five organ regions may introduce significant noise, thereby compromising the accuracy and specificity of the final radiology report. To address this issue, we evaluate the importance coefficient of each organ-regional cross-modal information. Furthermore, we leverage the strengths of GNNs in analyzing relationships among different items~(nodes)~\cite{gat}. The importance coefficient $\alpha_{\text{o}}$ for the organ-regional cross-modal information is defined as follows:
\begin{equation}
    \begin{aligned}
    &\alpha_{\text{o}} =  \text{MLP} \left( \sum_{u \in \mathcal{N}(v)} \mathbf{W}^{(k)} \mathbf{h}_u^{(k-1)} + \mathbf{b}^{(k)} \right),
    \end{aligned}
\end{equation}

where $W$ is the adjacent matrix generated in advance, which is obtained from knowledge graph~\cite{kiut}, $\mathcal{N}(v)$ is a set of the nodes $\mathcal{N}(v) \in \{x^{C}_{T}, x^{C}_{\text{o}}\}$. $\mathbf{h}_u^{k-1}$ is the value of the last layer's node $u$. $k$ is the layer number, $b$ is a learnable bias, and $\text{MLP}$ is the multilayer perceptron~\cite{tolstikhin2021mlp}.

Then, the final cross-modal feature $x^{C}_{F}$ can be obtained from the process:
\begin{equation}
    x^{C}_{F} = x^{C}_{\text{o}} \sum_{i\in \phi } x^{C}_{i} \alpha_{i},
\end{equation}
where $\phi$ is the organ set \{mediastinum, pleural, lung, heart,bone\}, $\oplus$ is the element-wise adding. Finally, we can get the final input image feature $x_{I} = x^{C}_{F} + x^{I_{r}}_{F}$, where $x^{I_{r}}_{F} = E_{r}(I_{r})$ is the fine-grained raw image feature extracted from the output layer of the image feature extractor $E_{r}$.

\subsection{Radiology Report Generation Module}
\label{subsection: Radiology Report Generation Module}
Following~\cite{R2GenCMN}, the final input image feature $x_{I}$ will input the encoder-decoder model~\cite{transformer} to generate the final radiology report. Specifically, the encoder model will first extract image features, which are used in the decoder model to generate the final radiology report. During the training process, we additionally introduce the consistency constraint loss to align the image feature~$\hat{x_{I}}$ after encoder and the radiology report $x_{T}$ after the embedding layer from the decoder, the loss function is defined as follows:
\begin{equation}
    \small
    \mathcal{L}_{CS} = 1 - \frac{\hat{x_{I}}^{\top} x_{T}}{||\hat{x_{I}}||~||x_{T}||}.
\end{equation}
The overall loss function can be defined as:
\begin{equation}
    \small
    \mathcal{L}_{T} =  \mathcal{L}_{CE} + \beta \times \mathcal{L}_{CS},
\end{equation}
where $\mathcal{L}_{CE}$ is the cross-entropy loss~\cite{mao2023cross} for report generation and $\mathcal{L}_{CS}$ is the cross-modal consistent loss with its coefficient $\beta$ commonly set as 0.1.

%% file: tables/instruction_tuning_example.tex
\begin{table}[thbp]
\centering
\resizebox{\linewidth}{!}{
\begin{tcolorbox}[width=0.5\textwidth, fontupper=\small, colback=blue!2, boxrule=0.9pt] 
Prompt: What have you found in \textcolor{purple}{$<$organ$>$}?\textbackslash n $<$image$>$\\
Answer: $<$organ-level diagnosis description$>$.

\end{tcolorbox}}
\captionof{figure}{Input and output type during the instruction tuning.}
\label{tab:prompt1} 
\vspace{-10px}
\end{table}

%% file: tables/SOTA_comparison_table.tex
\begin{table*}[!ht]
  \centering
  \begin{tabular}{c|l|ccc>{\columncolor{light_gray_ours}}ccc}
    \toprule
    \multicolumn{1}{c|}{\multirow{2}{*}{\textbf{\textsc{Dataset}}}} &
    \multicolumn{1}{c|}{\multirow{2}{*}{\textbf{\textsc{Method}}}} &
    \multicolumn{6}{c}{\textbf{\textsc{NLG Metric}}}
    \\\cmidrule(lr){3-8} ~ & ~ & BLUE@1 & BLUE@2 & BLUE@3 &BLUE@4 &METOR &ROUGE-L  \\ \midrule [\heavyrulewidth] \midrule
    $ $ &\textsc{DCL}~\cite{DCL}  & - & - & - & 0.163 & 0.193 & 0.383 \\
    $ $ & \textsc{MMTN}~\cite{MMTN}  & 0.486 & 0.321 & 0.232 & 0.175 & - & 0.375 \\
    \textbf{IU-} & \textsc{M2KT}~\cite{M2KT} & 0.497 & 0.319 & 0.230 & 0.174 & - & 0.399 \\
    \textbf{Xray} & \textsc{C2M-DOT}~\cite{C2M-DOT}  & 0.475 & 0.309 & 0.222 & 0.170 & 0.191 & 0.375 \\
    $ $ & \textsc{CMMRL}~\cite{CMM}  & 0.494 & 0.321 & 0.235 & 0.181 & 0.201 & 0.384 \\
    $ $ & \textsc{XPRONET$^{*}$}~\cite{Xpronet}  & 0.501 & 0.324 & 0.224 & 0.165 & 0.204 & 0.380 \\
    $ $ & \textsc{R2GenCMN$^{*}$}~\cite{R2GenCMMRL} & 0.475 & 0.309 & 0.222 & 0.165 & 0.187 & 0.371 \\
    \cmidrule(lr){2-8}
  $ $ & \textsc{ORID(Ours)} & \textbf{0.501} & \textbf{0.351} & \textbf{0.261} & \textbf{0.198} & \textbf{0.211} & \textbf{0.400} \\
      \midrule 
      \midrule 
    $ $ & \textsc{DCL}~\cite{DCL} & - &  - &  - &  0.109 &  0.150 & 0.284  \\
    $ $ & \textsc{MMTN}~\cite{MMTN} & 0.379 & 0.238 & 0.159 & 0.116 & \textbf{0.160} & 0.283 \\
    \textbf{MIMIC} & \textsc{M2KT}~\cite{M2KT} &  0.386 & 0.237 & 0.157 &  0.111 &  - & 0.274 \\
    \textbf{CXR} & \textsc{Lgi-MIMIC}~\cite{Longitudinal-MIMIC} & 0.343 & 0.210 & 0.140 & 0.099 & 0.137& 0.271 \\
    $ $ & \textsc{CMMRL}~\cite{CMM} & 0.353 & 0.218 & 0.148 & 0.106 & 0.142 & 0.278  \\
    $ $ & \textsc{XPRONET}~\cite{Xpronet} & 0.344 & 0.215 & 0.146 & 0.105 & 0.138 & 0.279 \\
    $ $ & \textsc{R2GenCMN$^{*}$}~\cite{R2GenCMMRL} & 0.347 &  0.221 & 0.139 & 0.097 & 0.138 & 0.274 \\
    \cmidrule(lr){2-8}
    $ $ & \textsc{ORID(Ours)} & \textbf{0.386} & \textbf{0.238} & \textbf{0.163} & \textbf{0.117} & 0.150 & \textbf{0.284} \\
    \bottomrule
  \end{tabular}
  \caption{The results of the ORID model and other tested models in IU-Xray and MIMIC-CXR benchmarks. $*$ indicates we reproduced. The results for other models are obtained from their original papers. The best result is presented in \textbf{bold}. The most important metric has been marked in \colorbox{light_gray_ours}{grey}.}
  \label{table: result compare}
  \vspace{-0.4cm}
\end{table*}

%% file: sections/experiments_setting.tex
\section{Experiment Settings}
\label{sec:experiment setting}

\subsection{Datasets}
\label{subsection: datasets}
To rigorously evaluate the efficacy of our proposed method, we conduct experiments on two widely recognized public benchmarks: IU-Xray~\cite{iu_xray} and MIMIC-CXR~\cite{mimic_cxr}. The IU-Xray dataset\footnote{\url{https://openi.nlm.nih.gov/faq}} containing 7,470 chest X-ray images and 3,955 corresponding reports. In contrast, the MIMIC-CXR dataset\footnote{\url{https://physionet.org/content/mimic-cxr/2.0.0/}}, is significantly larger and comprising 473,057 chest X-ray images and 206,563 associated reports. Consistent with the methodology outlined in \cite{R2GenCMN}, both datasets were partitioned into training, validation, and testing sets in a 7:2:1 ratio.

\subsection{Baseline and Evaluation Metrics}
\label{subsection: baselien and evaltion metrics}

\paragraph{Baseline.} 
To rigorously assess the performance of our ORID framework in the radiology report generation task, we conduct a comparative analysis against several state-of-the-art models. Specifically, we compare our method with DCL~\cite{DCL}, MMTN~\cite{MMTN}, M2KT~\cite{M2KT}, C2M-DOT~\cite{C2M-DOT}, Lgi-MIMIC~\cite{Longitudinal-MIMIC}, CMMRL~\cite{CMM}, R2GenCMN~\cite{R2GenCMN}, and XRONET~\cite{Xpronet}. Certain models have been excluded from our evaluation for well-defined reasons. The RGRG~\cite{rgrg} has been omitted due to its significantly large architecture, incorporating a 24-layer decoder, which presents substantial challenges for reproducibility among other researchers. Similarly, the RepsNet~\cite{repsnet} and METransformer~\cite{METransformer} have been excluded as they employ a distinct dataset split for evaluation and testing. Moreover, the KiUT~\cite{kiut} has been excluded due to the unavailability of its source code, impeding the reproducibility of its results. This rigorous selection process ensures equitable and consistent comparisons, thereby bolstering the reliability of our evaluation and the validity of our conclusions.

\paragraph{Evaluation Metrics.}
In this paper, we employ Natural Language Generation~(NLG) metrics such as BLEU~\cite{bleu}, ROUGE-L~\cite{rougel}, and METEOR~\cite{meteor} to evaluate our model. BLEU measures word n-gram overlap, ROUGE-L assesses sentence-level coherence via the longest common subsequence, and METEOR evaluates text similarity with precision, recall, and semantic analysis. These metrics robustly assess the accuracy and coherence of generated radiology reports. Additionally, following previous works~\cite{R2GenCMN, R2GenCMMRL}, we employ clinical efficacy metrics to evaluate the quality of the generated reports on the MIMIC-CXR dataset~\cite{mimic_cxr}. These metrics can assess the presence of a predefined set of clinically significant observations, thereby providing a measure of the diagnostic accuracy and relevance of the generated reports compared to the corresponding target reports.

\subsection{Implement Details}
\label{subsection: implement details}
In our approach, we utilize the ResNet101~\cite{resnet} and ResNet18~\cite{resnet} as the image feature extractor and mask feature extractor, which are initialized with pretrained weights from ImageNet. The CXAS~\cite{cxas} model is used to segment organ masks from radiology images. We adopt the graph attention network~\cite{gat} with eight heads to analyze relationships between different organs. We train LLaVA-Med-RRG and our proposed ORID framework on 4 and 1 NVIDIA A100 GPUs (80G), respectively. We employ the Adam~\cite{adam} optimizer initialized with a learning rate of $1e-4$ for the image extractor and $5e-4$ for the other components. The encoder-decoder in the radiology report generation module is trained using the teacher forcing method~\cite{lamb2016professor}. During inference, we utilize beam search~\cite{beam_search} with a width of 3. We train our model for 100 and 30 epochs on the IU-Xray~\cite{iu_xray} and MIMIC-CXR~\cite{mimic_cxr} datasets. To enhance model robustness, images and masks underwent data augmentation techniques, including random cropping and random horizontal flipping. For both benchmarks, the numbers of masks for the heart, lung, bone, pleural, and mediastinum are 6, 15, 70, 10, and 9. The diagnosis description token lengths for the above five organs are 39, 53, 48, 43, and 41. Additionally, the final radiology reports associated with the datasets are preprocessed, involving the removal of punctuation and the conversion of infrequent words~(appearing less than three times) to the $\mathtt{[UNK]}$ token.

%% file: sections/experiment_result_and_comparison.tex
\section{Experiments}
\label{sec:experiment results and comparison}

\input{tables/CE_metrics_compare}
\subsection{Comparison with State-of-the-art models}
\label{subsection: Comparison with state-of-the-art}

\definecolor{red_fig4}{rgb}{1.0, 0.0, 0.09}
\definecolor{green_fig4}{rgb}{0.2, 0.62, 0.21}
\begin{figure*}[ht]
\begin{center}
\includegraphics[width=1.0\linewidth]{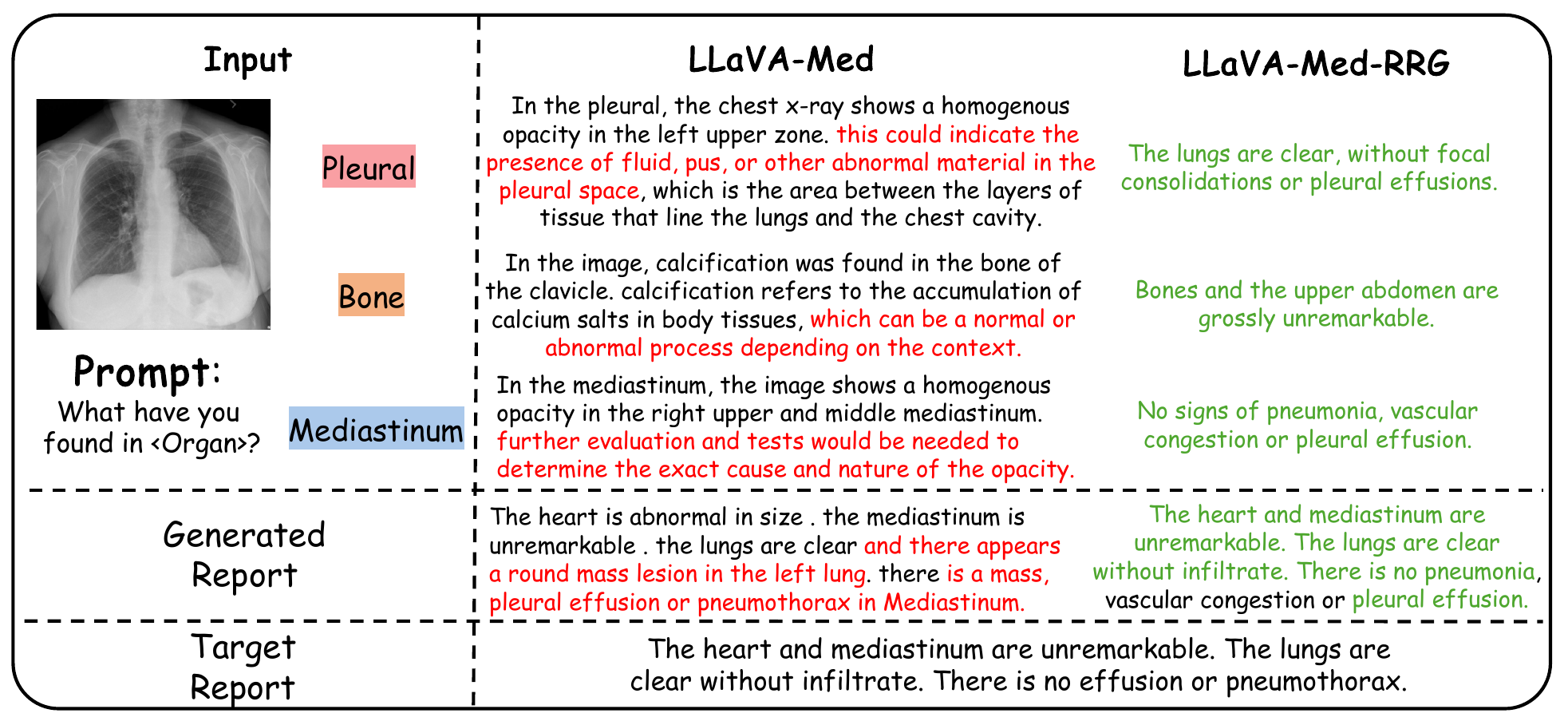}
\end{center}
\vspace{-0.4cm}
\caption{An example of LLaVA-Med's organ-reional diagnosis description compare with that of LLaVA-Med-RRG. The sentences that are correct or highly-related with target reports have been marked in {\color{green_ours}{green}}, otherwise have been marked in {\color{red_fig4}{red}}.}
\label{fig: the example of LLaVA-Med's organ-level diagnosis description}
\vspace{-0.4cm}
\end{figure*}

\paragraph{Descriptive Accuracy.} The results of our descriptive accuracy analysis are presented in Tab.~\ref{table: result compare}. Our proposed ORID framework demonstrates superior performance compared to the SOTA models across all assessed metrics on both datasets. Notably, on the IU-Xray dataset, ORID excels in all metrics. Similarly, on the MIMIC-CXR dataset, ORID showcases superior performance in all BLEU metrics and ROUGE-L. It can be noticed that our method obtains a slightly lower METEOR score than MMTN~\cite{MMTN}, this can be attributed to our framework focusing on disease detection and analysis, which leads to a marginal reduction in report diversity. 

\begin{figure}[!t]
\begin{center}
\includegraphics[width=1.0\linewidth]{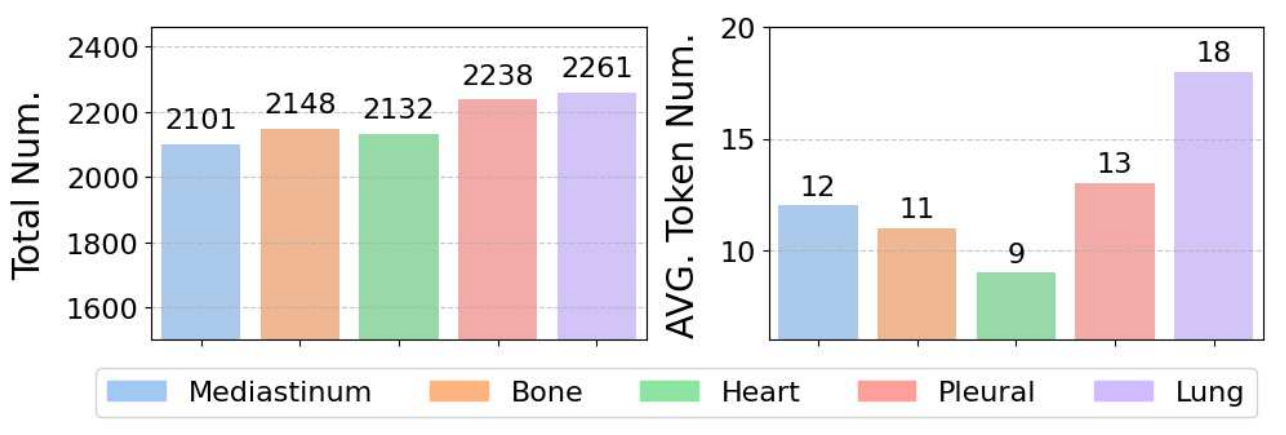}
\end{center}
\vspace{-0.2cm}
\caption{Statistical analysis of question-answer pairs and average token length for each organ.}
\label{fig: instruction data analysis table chart}
\vspace{-0.2cm} 
\end{figure}

\input{tables/word_cloud_image}

\paragraph{Clinical Correctness.} 
In Tab.~\ref{table:comp_CE_Metric}, we present a comparative analysis of our proposed ORID model against four SOTA models: R2Gen~\cite{R2Gen}, CMMRL~\cite{R2GenCMMRL}, METransformer~\cite{METransformer}, and R2GenCMN~\cite{R2GenCMN}, on the MIMIC-CXR dataset~\cite{mimic_cxr}. The experiment results show that our model demonstrates superior performance in all the metrics. Specifically, compared with the METransformer, our model achieves a 19.5\% improvement in precision and a 13.2\% improvement in F1-score. This can be attributed to our framework fully utilizing the organ-regional information which significantly improves the capability in organ-level disease analysis. Besides, our framework obtains competitive results in Recall compared to the METransformer, which utilizes multiple experts that excel in detecting a broad range of diseases.

\subsection{Ablation Study and Analysis}
\input{tables/ablation_instruction_tuning}

\paragraph{Analysis on LLaVA-Med-RRG.}
The LLaVA-Med-RRG is instruction tunned based on it by our constructed organ-level RRG-related instruction dataset, which comprises approximately 10k QA pairs derived from nearly 4K radiology image-report pairs. Specifically, as illustrated in Fig.~\ref{fig: instruction data analysis table chart}, the dataset includes 2.2k, 2.1k, 2.2k, 2.1k, and 2.1k pairs for the {\colorbox{pink_ours}{pleural}}, {\colorbox{green_ours}{heart}}, {\colorbox{purple_ours}{lung}}, {\colorbox{orange_ours}{bone}}, and {\colorbox{blue_ours}{mediastinum}}, respectively. All the organ-regional diagnosis descriptions are concise and contain fewer than 20 tokens. Furthermore, we provide a word cloud analysis for different organs and the whole dataset in Fig.~\ref{fig: instruction data analysis world cloud}. This analysis demonstrates substantial QA diversity for each organ, with a predominant focus on disease-related queries (such as pleural effusion, rib fracture, and pulmonary edema) and a lesser emphasis on normal condition analysis. Notably, most regional diagnostic descriptions focus on heart size due to the relative scarcity of heart diseases.

We conduct a comparison experiment between LLaVA-Med-RRG and LLaVA-Med in Tab.~\ref{table: Ablation study on Instruction tuning dataset}. Experiment results indicate that LLaVA-Med-RRG significantly outperforms LLaVA-Med in the downstream RRG task across all evaluated metrics. Fig.~\ref{fig: the example of LLaVA-Med's organ-level diagnosis description} provides a visual comparison of organ-regional descriptions generated by LLaVA-Med and our model. The figure demonstrates that the diagnosis descriptions generated by LLaVA-Med are excessively lengthy and often introduce noise and inaccuracies. These issues negatively affect the final generated reports, leading to the inclusion of errors and inefficient use of computational resources due to the processing of long tokens. In contrast, the descriptions produced by our model, LLaVA-Med-RRG, are concise and more relevant, resulting in more accurate and efficient radiology report generation.

\input{tables/ablation_study_table}
\vspace{-0.03cm}
\label{subsection: ablation study}
\paragraph{Ablation on Different Modules.}
Considering the extensive size of the MIMIC-CXR dataset and our objective to minimize CO2 emissions, we conduct an ablation study on the IU-Xray dataset to assess the influence of each module in our approach. In Tab.~\ref{table: ablation study}, BL denotes the baseline model, OCF refers to the Organ-based Cross-modal Information Module~(Sec.~\ref{subsection: Global-Local Crosss-modal Feature Fusion Module}), F and C represent fine-grained and coarse-grained analyses respectively, and OICA signifies the Organ Importance Coefficient Analysis Module~(Sec.~\ref{subsection: Organ Importance Coefficient Analysis Module}). A comparison between configurations \#1 and \#2 in Tab.~\ref{table: ablation study} indicates that incorporating the organ mask leads to moderate enhancements in the model's RRG capability. This marginal improvement can be attributed to the challenges in distinguishing abnormal regions and disease detection when only organ-related regions are highlighted in the image. However, the inclusion of the OCF module significantly enhances caption accuracy, as evidenced by the comparison between \#2 and \#3. The fusion of cross-modal features from organ-specific diagnostic descriptions and image features notably boosts the accuracy of final reports across all evaluation metrics. Furthermore, comparing \#3 and \#4 reveals additional benefits from combining coarse and fine-grained analyses. Ultimately, \#5, encompassing all proposed contributions demonstrates substantial performance gains across various metrics.

\begin{figure}[!t]   
\begin{center}
\includegraphics[width=1\linewidth]{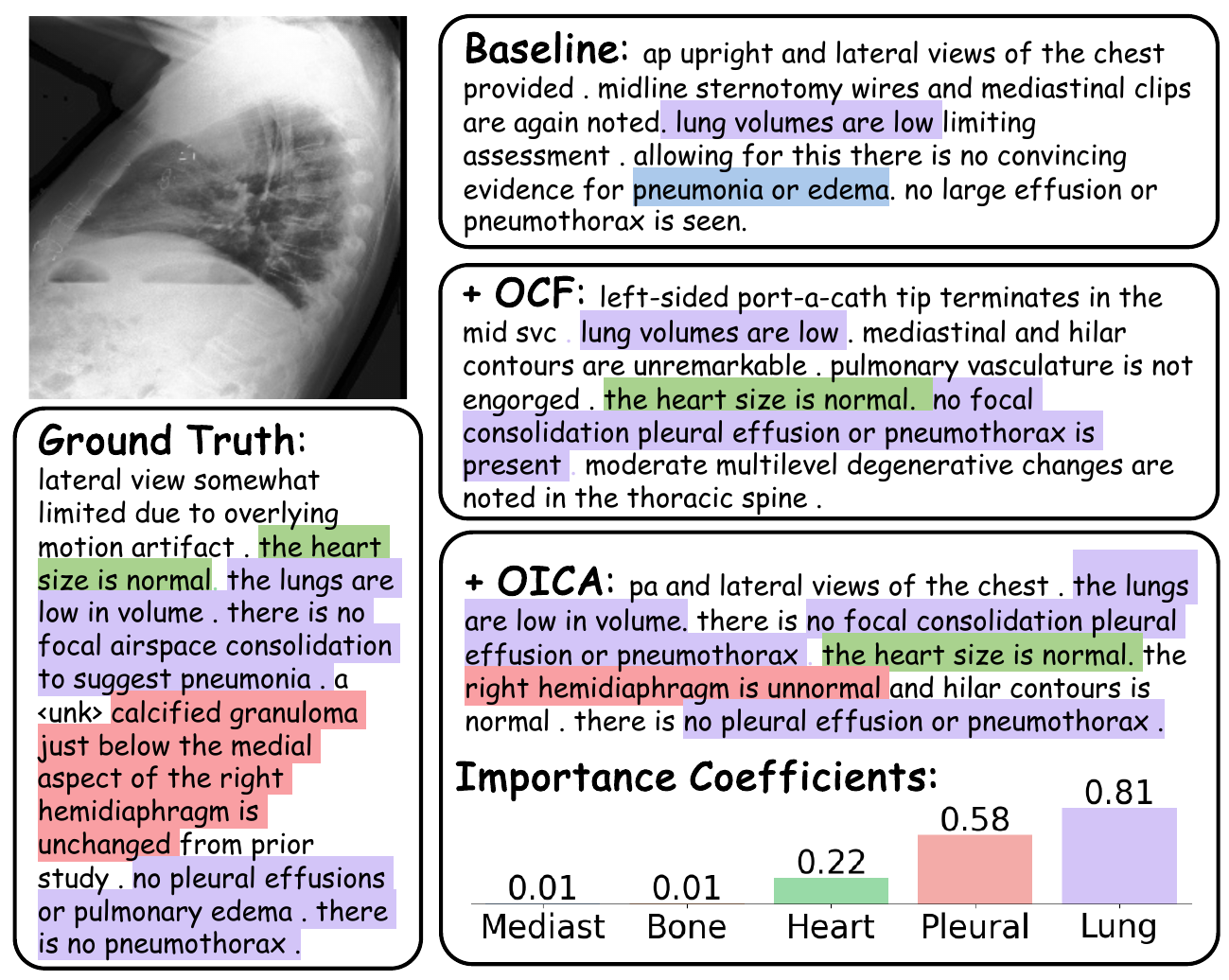}
\end{center}
\vspace{-0.3cm}
\caption{Qualitative examples of generated radiology reports with different modules.}
\vspace{-0.3cm}
\label{Fig: the example of visualization}
\end{figure}

\subsection{Qualitative Analysis}
\label{subsection: qualitative analysis}
To comprehensively evaluate the effectiveness of the ORID framework, we conduct a qualitative analysis based on the MIMIC-CXR dataset. The qualitative examples are illustrated in Fig.~\ref{Fig: the example of visualization}, where we specifically highlight the {\colorbox{pink_ours}{pleural}}, {\colorbox{green_ours}{heart}}, {\colorbox{purple_ours}{lung}}, {\colorbox{orange_ours}{bone}}, and {\colorbox{blue_ours}{mediastinum}} with different colours in the radiology report. As demonstrated in Fig.~\ref{Fig: the example of visualization}, compared with the baseline, the integration of the OCF module leads to reports containing more pertinent information than the baseline, including specifics on lung diseases, better aligning with the desired report content. Furthermore, the addition of the OICA module enhances the final report by incorporating comprehensive information about the heart. Moreover, we present the importance coefficients for the five organ regions in Fig.~\ref{Fig: the example of visualization}, showcasing that the heart, pleura, and lung achieve the highest scores. As a result, the final report primarily emphasizes these three organ regions, reflecting the emphasis seen in the target report.

%% file: tables/CE_metrics_compare.tex
\begin{table}[!t]
\resizebox{\linewidth}{!}{
\begin{tabular}{l|cc>{\columncolor{light_gray_ours}}c}
\toprule
\multirow{2}{*}{\textbf{\textsc{Method}}} & \multicolumn{3}{c}{\textbf{\textsc{CE Metric}}} \\
\cmidrule(lr){2-4} & Precision & Recall & F1-Score \\
\midrule [\heavyrulewidth] \midrule
\textsc{R2Gen} \cite{R2Gen}              & 0.333  & 0.273 & 0.276 \\
\textsc{CMMRL} \cite{R2GenCMMRL}       & 0.342  & 0.294 & 0.292 \\
\textsc{R2GenCMN} \cite{R2GenCMN}       & 0.334  & 0.275 & 0.278 \\
\textsc{METransformer} \cite{METransformer}  & 0.364  & \textbf{0.309} & 0.311 \\
\midrule
\textsc{ORID(Ours)}    & \textbf{0.435}  &  0.295   &  \textbf{0.352}       \\
\bottomrule
\end{tabular}
}
\label{table: CE_metrics_compare}
\caption{Comparison of clinical efficacy metrics for the MIMIC-CXR dataset. The best result is presented in \textbf{bold}. The critical metrics have been shaded in \colorbox{light_gray_ours}{grey}.}
\label{table:comp_CE_Metric}
\vspace{-0.4cm}
\end{table}

%% file: tables/word_cloud_image.tex
\begin{figure}[!t]
    \centering
    \begin{minipage}{0.15\textwidth}
        \centering
        \includegraphics[width=\textwidth]{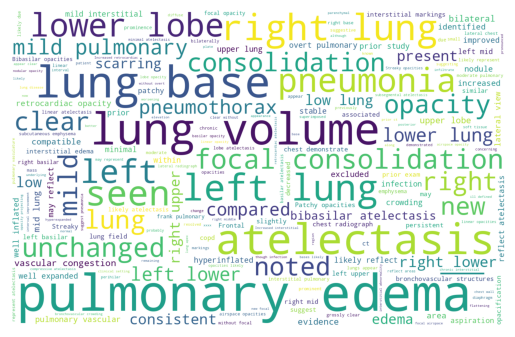}
        \subcaption{Lung}
    \end{minipage}
    \begin{minipage}{0.15\textwidth}
        \centering
        \includegraphics[width=\textwidth]{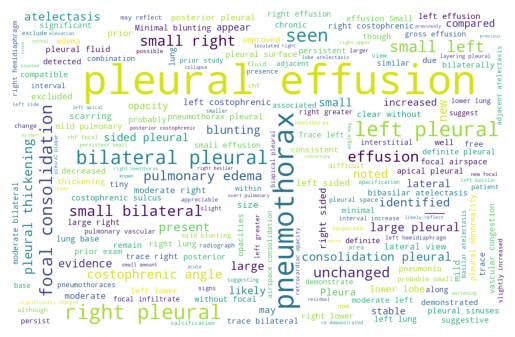}
        \subcaption{Pleural}
    \end{minipage}
    \begin{minipage}{0.15\textwidth}
        \centering
        \includegraphics[width=\textwidth]{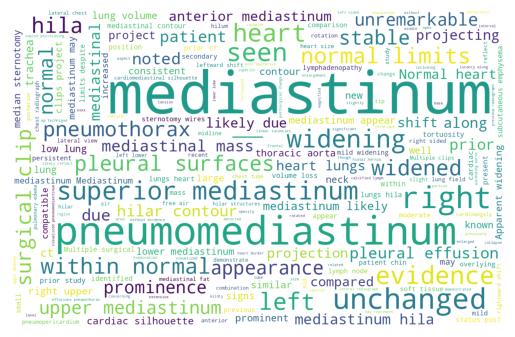}
        \subcaption{Mediastinum}
    \end{minipage}


    \begin{minipage}{0.15\textwidth}
        \centering
        \includegraphics[width=\textwidth]{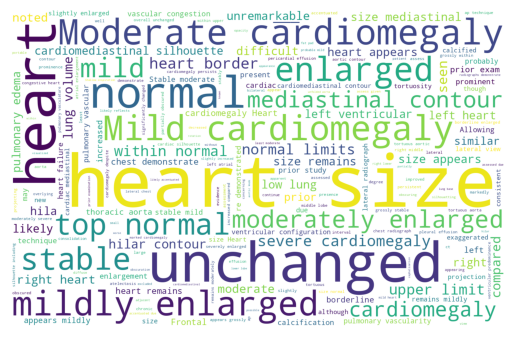}
        \subcaption{Heart}
    \end{minipage}
    \begin{minipage}{0.15\textwidth}
        \centering
        \includegraphics[width=\textwidth]{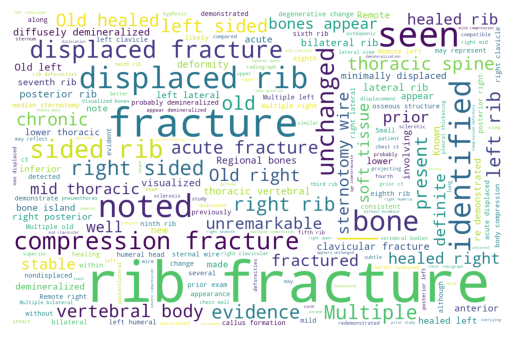}
        \subcaption{Bone}
    \end{minipage}
    \begin{minipage}{0.15\textwidth}
        \centering
        \includegraphics[width=\textwidth]{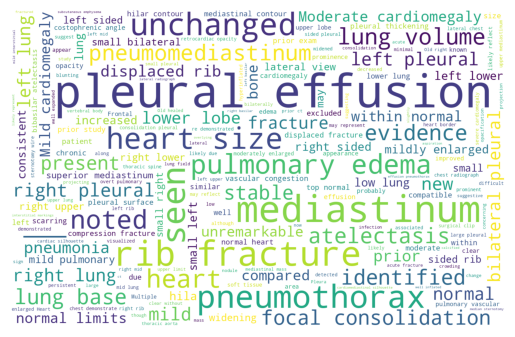}
        \subcaption{Total}
    \end{minipage}
    \caption{The word cloud analysis about each organ and total in instruction-tuning dataset.}
    \label{fig: instruction data analysis world cloud}
    \vspace{-0.3cm}
\end{figure}

%% file: tables/ablation_instruction_tuning.tex
\begin{table}[!t]
  \centering
  \begin{tabular}{l|c>{\columncolor{light_gray_ours}}ccc}
    \toprule
    \textbf{\textsc{Diagnosis Model}} & B@1 & B@4 &MTR. & RGL.  \\ \midrule [\heavyrulewidth] \midrule
    \textsc{LLaVA-Med}~\cite{llava-med} & 0.441  & 0.158 & 0.179 & 0.378 \\
    \textsc{LLaVA-Med-RRG} & \textbf{0.501}  & \textbf{0.198} & \textbf{0.211} & \textbf{0.400} \\
    \bottomrule
  \end{tabular}
  \caption{Experiment comparison between LLaVA-Med-RRG and LLaVA-Med. The best result is presented in \textbf{bold}. The most important metric is marked in \colorbox{light_gray_ours}{grey}.}
  \label{table: Ablation study on Instruction tuning dataset}
  \vspace{-0.3cm}
\end{table}

%% file: tables/ablation_study_table.tex
\begin{table}[!t]
\centering
\footnotesize
\setlength{\tabcolsep}{4.5pt}
\begin{center}
\begin{tabular}{@{}c|c c c c c   |c >{\columncolor{light_gray_ours}}c c c@{}}
\toprule
\multirow{2}{*}[-3pt]{\textbf{\textsc{\#}}} & \multirow{2}{*}[-3pt]{\textbf{\textsc{BL.}}}  &\multirow{2}{*}[-3pt]{\textbf{\textsc{Mask}}} & \multicolumn{2}{c}{\textbf{\textsc{OCF}}}  & \multirow{2}{*}[-3pt]{\textbf{\textsc{OICA}}} &
\multicolumn{4}{c}{\textbf{\textsc{Dataset: IU-Xray }}~\cite{iu_xray}}     \\
\cmidrule(lr){4-5} \cmidrule(lr){7-10}
& & & \textbf{\textsc{F}} & \textbf{\textsc{C}} & & \textsc{B@1} & \textsc{B@4} & MTR. & RGL. \\ \midrule [\heavyrulewidth]
\midrule
$ $ 1 & \checkmark & &  &  &  &0.475&  0.165  & 0.187  & 0.371  \\
$ $ 2  & \checkmark & \checkmark  & &  &  & 0.498 & 0.159 & 0.187 & 0.374 \\
$ $ 3 &\checkmark  & \checkmark  &\checkmark  & &  & 0.501 &   0.170 &  0.206 & 0.360 \\ 
$ $ 4 &\checkmark  & \checkmark  &\checkmark  & \checkmark &  & \textbf{0.503} &   0.172 &  0.211 & 0.354 \\ 
$ $ 5 &\checkmark  & \checkmark &\checkmark  &\checkmark  &\checkmark  & 0.501 & \textbf{0.198} & \textbf{0.211} & \textbf{0.400}   \\
\bottomrule
\end{tabular}
\end{center}
\caption{Ablation study on different modules of ORID. The best result is presented in \textbf{bold}. The most important metric is marked in \colorbox{light_gray_ours}{grey}.}
\label{table: ablation study}
\vspace{-0.3cm}
\end{table}

%% file: sections/conclusion.tex
\section{Conclusion}
\label{sec:conclusion}
This paper presents a novel Organ-Regional Information
Driven~(ORID) framework for generating accurate and believable radiology reports. Initially, leveraging LLaVA-Med, we establish an RRG-oriented instruction dataset to enhance organ-regional diagnosis descriptions and obtain LLaVA-Med-RRG. Subsequently, we introduce an organ-based cross-modal fusion module to effectively integrate information from organ-specific diagnosis descriptions and radiology images. To mitigate the impact of irrelevant organ noise on radiology report generation, we propose an organ importance coefficient analysis module employing graph neural networks to analyze cross-modal information interconnections within each organ region. Our framework shows superior performance across various evaluation metrics in different benchmarks. We hope that our work provides insights into the radiology report generation field.

%% file: sections/appendix.tex
\appendix

\section{Appendix}

\subsection{Disease Symptom Graph}

\begin{figure}[!ht]   
\begin{center}
\includegraphics[width=1\linewidth]{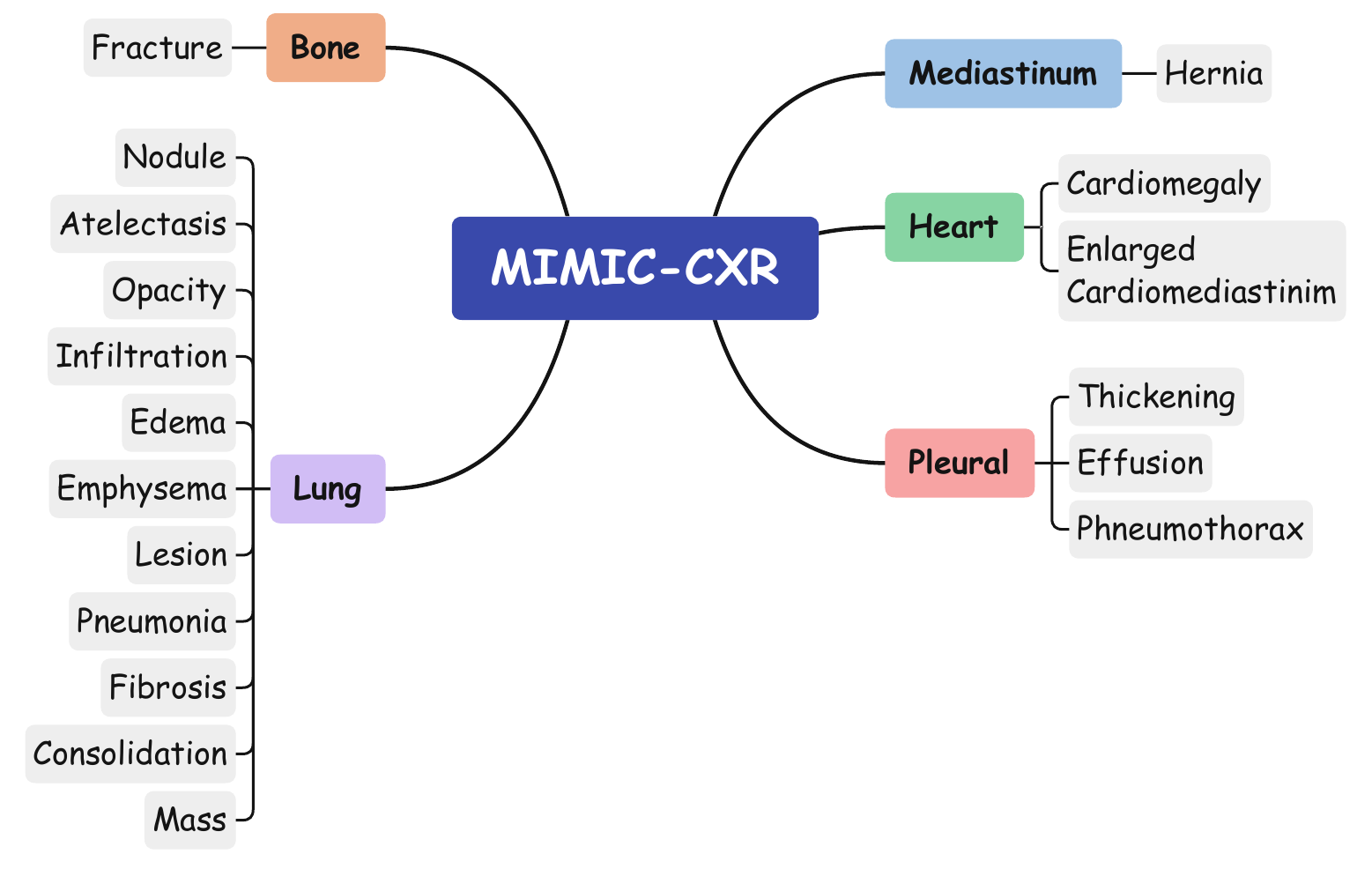}
\end{center}
\vspace{-0.5cm}
\caption{The symptom graph summarizes the related diseases for each organ in the MIMIC-CXR dataset.}
\label{Fig: disease symptom graph}
\end{figure}

Fig.~\ref{Fig: disease symptom graph} illustrates the detailed knowledge graph of disease symptoms derived from prior disease captions. This graph, referenced from \cite{kiut}, was constructed based on a professional analysis of the relationships between organs and their corresponding diseases as observed in radiology images. Utilizing this graph, we developed the instruction-tuning dataset and the adjacency matrix for the Graph Neural Network~(GNN).

\subsection{Benchmark Information}
\input{tables/supp_benchmark_information}

Table~\ref{table: supp bencmark dataset} presents comprehensive information on the two benchmark datasets employed to evaluate our ORID framework. The data indicate that the MIMIC-CXR dataset encompasses a greater number of cases compared to the IU-Xray dataset.

\subsection{Mask Information}
\input{tables/supp_regional_mask_images}

\begin{figure}[!ht]   
\begin{center}
\includegraphics[width=1\linewidth]{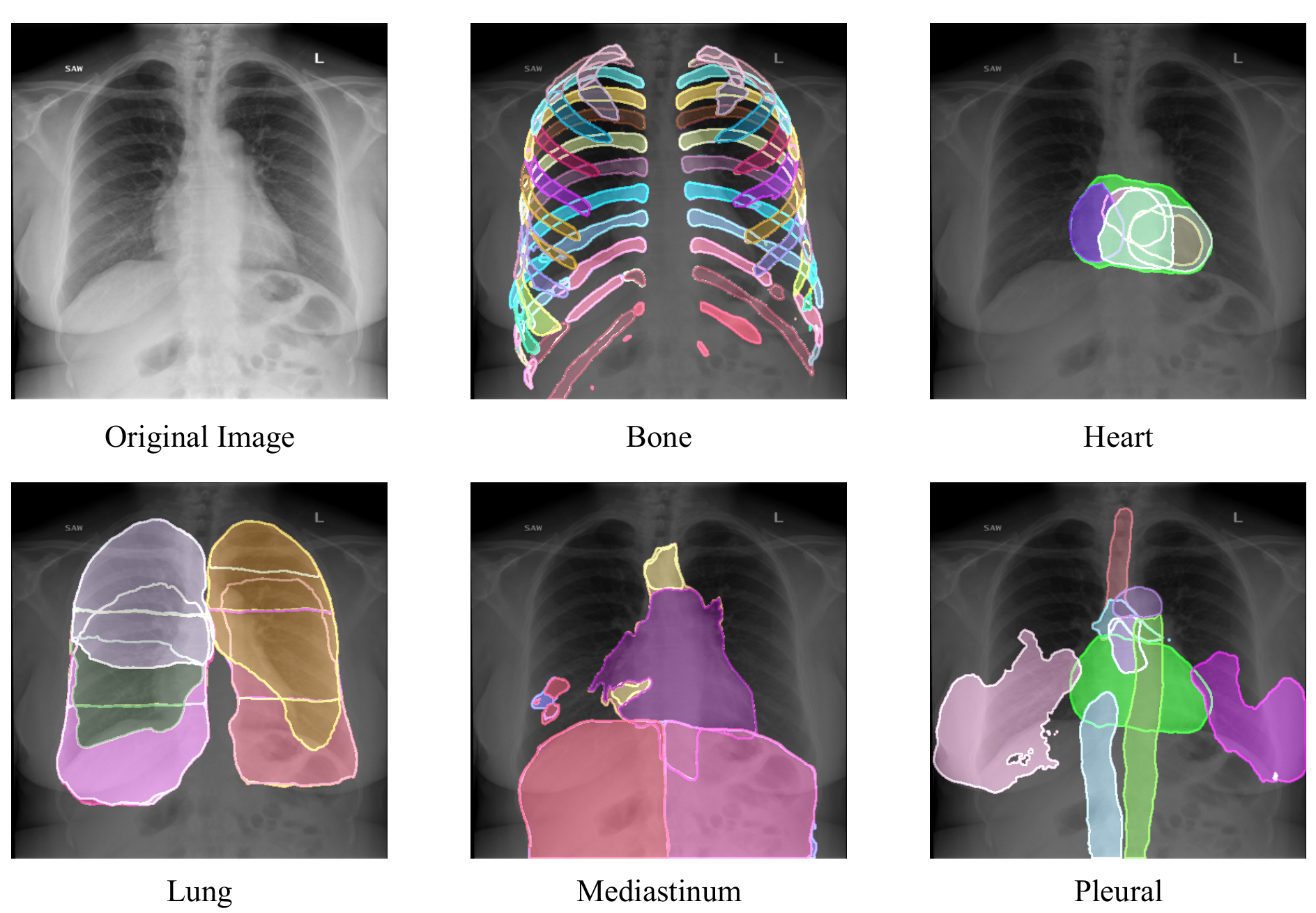}
\end{center}
\vspace{-0.5cm}
\caption{The visualization of the organ mask sets with the original image. Due to each organ region corresponding to several small organ parts, the different color means different part organ mask images in its corresponding regions.}
\label{Fig: The visualization of mask}
\end{figure}

Table~\ref{Appendix:table_mask_info} provides specific details regarding the small organ masks included for each organ set. Furthermore, the visualization of these mask sets alongside the original images is presented in Fig.~\ref{Fig: The visualization of mask}.

\subsection{Instruction-tuning Dataset}

\begin{figure*}[!ht]
    \centering
    \includegraphics[width=15cm]{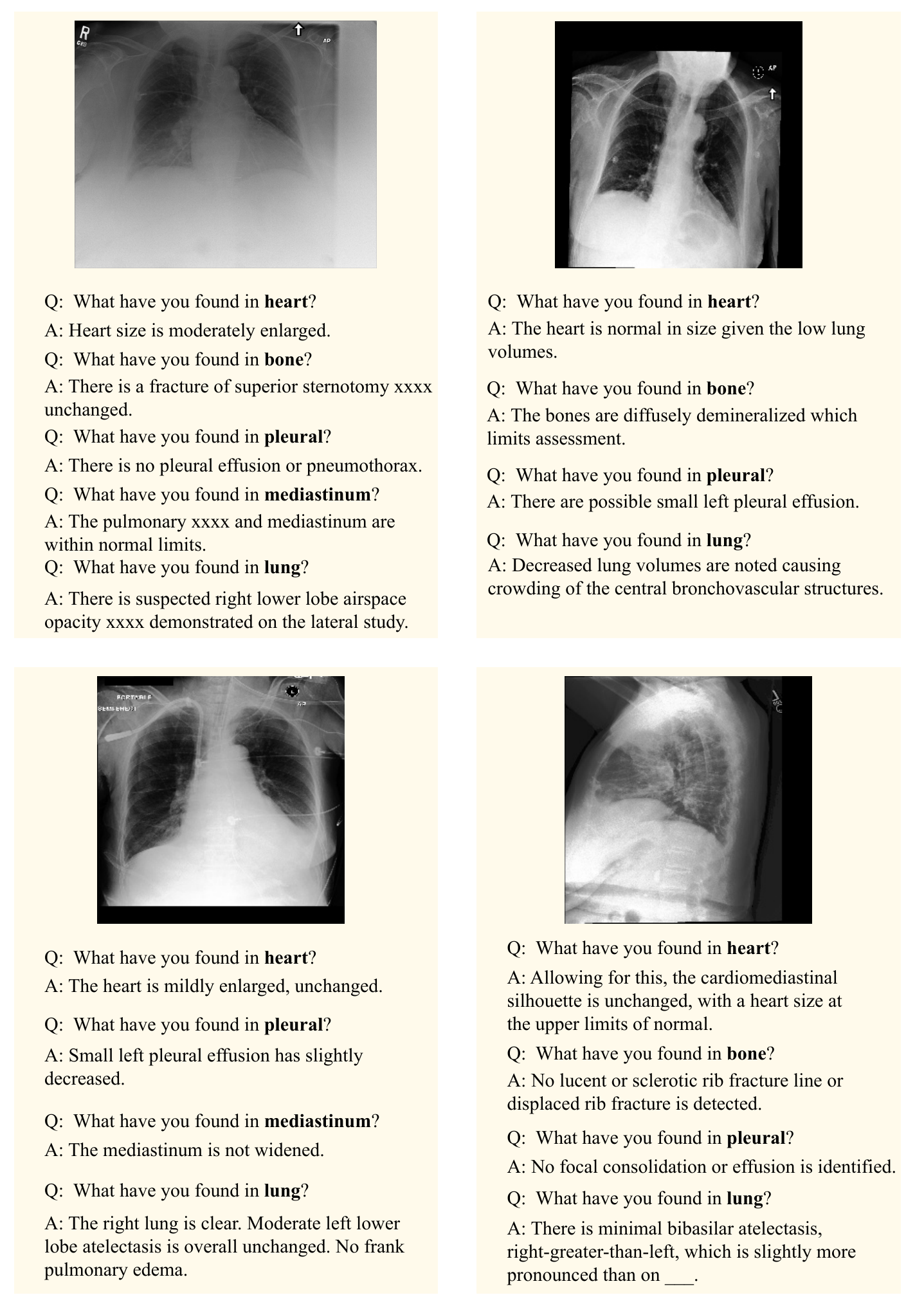}
    \caption{The examples about the RRG instruction dataset to instruction tuning the LLaVA-Med}
    \label{Fig: The example of rrg instruction dataset}
\end{figure*}

Fig.~\ref{Fig: The example of rrg instruction dataset} presents examples from the instruction-tuning dataset. Notably, each image is accompanied by more than four question-answer pairs pertaining to various organs.

\subsection{Case Study}
\begin{figure*}[ht]   
\begin{center}
\includegraphics[width=15cm]{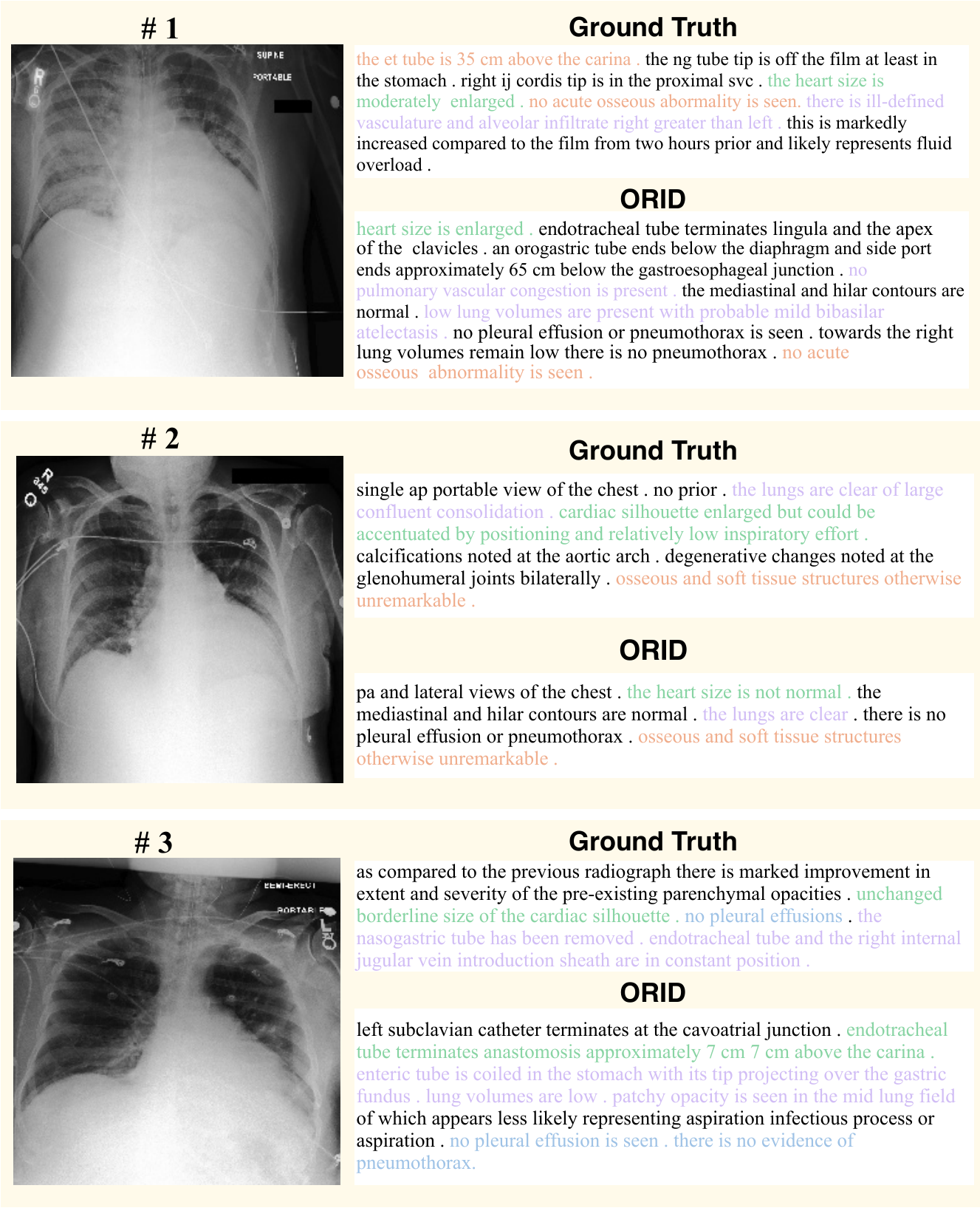}
\end{center}
\vspace{-0.5cm}
\caption{The visualization of prediction results by the ORID model. We specifically highlight the {\textcolor{pink_ours}{pleural}}, {\textcolor{green_ours}{heart}}, {\textcolor{purple_ours}{lung}}, {\textcolor{orange_ours}{bone}}, and {\textcolor{blue_ours}{mediastinum}} with different colours in the radiology report.}
\label{Fig: case study}
\end{figure*}

We have shown the results of our ORID framework generated compared with that of ground truth in Fig.~\ref{Fig: case study}. We have also marked the {\textcolor{pink_ours}{pleural}}, {\textcolor{green_ours}{heart}}, {\textcolor{purple_ours}{lung}}, {\textcolor{orange_ours}{bone}}, and {\textcolor{blue_ours}{mediastinum}} in different colors. More specifically, Example 1 shows the disease symptoms related to the heart and lungs; Example 2 shows the disease symptoms related to the heart.

%% file: tables/supp_benchmark_information.tex
\begin{table}[!ht]
\resizebox{\linewidth}{!}{
\begin{tabular}{l|lll|lll}
\toprule
\multirow{2}{*}{\text{\textsc{Dataset}}} & \multicolumn{3}{c|}{\text{\textsc{IU-Xray}}~\cite{iu_xray}} & \multicolumn{3}{c}{\text{\textsc{MIMIC-CXR}}~\cite{mimic_cxr}} \\
\cmidrule{2-7}              & \multicolumn{1}{c}{Train}   & \multicolumn{1}{c}{Val.}       & \multicolumn{1}{c|}{Test}  & \multicolumn{1}{c}{Train}    & \multicolumn{1}{c}{Val.}       & \multicolumn{1}{c}{Test}  \\
\midrule
\midrule
\text{\textsc{Image}}                    & 5.2K    & 0.7K       & 1.5K  & 369.0K   & 3.0K       & 5.2K  \\
\text{\textsc{Report}}                   & 2.8K    & 0.4K       & 0.8K  & 222.8K   & 1.8K       & 3.3K  \\
\text{\textsc{Patient}}                  & 2.8K    & 0.4K       & 0.8K  & 64.6K    & 0.5K       & 0.3K  \\
\text{\textsc{Avg. Len.}}                & 37.6    & 36.8       & 33.6  & 53.0     & 53.1       & 66.4  \\
                        \bottomrule
\end{tabular}
}
\caption{The specifications of two benchmark datasets that will be utilized to test the ORID model.}
\label{table: supp bencmark dataset}
\end{table}

%% file: tables/supp_regional_mask_images.tex
\begin{table}[!ht]
\resizebox{\linewidth}{!}{
\begin{tabular}{llcclll}
\toprule
\multicolumn{1}{l|}{\textsc{Organ Mask}} & \multicolumn{1}{l|}{\textsc{Num.}} & \multicolumn{1}{l|}{\textsc{Region}} & \multicolumn{1}{l}{\textsc{Total Mask}} \\ 
\midrule \midrule
\multicolumn{1}{l|}{Lung lobes}    & \multicolumn{1}{l|}{5}  & \multicolumn{1}{c|}{\multirow{3}{*}{Lung}} & \multirow{12}{*}{159} \\
\multicolumn{1}{l|}{Lung zones}    & \multicolumn{1}{l|}{8}  & \multicolumn{1}{c|}{}                      &      \\
\multicolumn{1}{l|}{Lung halves}   & \multicolumn{1}{l|}{2}  & \multicolumn{1}{c|}{}                      &      \\ \cmidrule{1-3}
\multicolumn{1}{l|}{Heart region}  & \multicolumn{1}{l|}{6}  & \multicolumn{1}{c|}{Heart}                 &      \\ \cmidrule{1-3}
\multicolumn{1}{l|}{Mediastinum}   & \multicolumn{1}{l|}{6}  & \multicolumn{1}{c|}{\multirow{2}{*}{Mediastinum}}   &  \\
\multicolumn{1}{l|}{Diaphragm}   & \multicolumn{1}{l|}{3}  & \multicolumn{1}{c|}{}   &  \\ \cmidrule{1-3}
\multicolumn{1}{l|}{Ribs}          & \multicolumn{1}{l|}{46} & \multicolumn{1}{c|}{\multirow{2}{*}{Bone}} &    \\
\multicolumn{1}{l|}{Ribs super}    & \multicolumn{1}{l|}{24} & \multicolumn{1}{c|}{}                      &           \\ \cmidrule{1-3}
\multicolumn{1}{l|}{Trachea}          & \multicolumn{1}{l|}{2} & \multicolumn{1}{c|}{\multirow{3}{*}{Pleural}} &    \\
\multicolumn{1}{l|}{Vessels}    & \multicolumn{1}{l|}{6} & \multicolumn{1}{c|}{}                      &           \\
\multicolumn{1}{l|}{Breast Tissue}    & \multicolumn{1}{l|}{2} & \multicolumn{1}{c|}{}                      &           \\ \cmidrule{1-3}
\multicolumn{1}{l|}{...}           & \multicolumn{1}{l|}{...}& \multicolumn{1}{l|}{...}                   & \multicolumn{1}{l}{} \\
\bottomrule
\end{tabular}
}
\caption{The specific information of masks generated by the CXAS model \cite{cxas}, as well as the mask images we ultimately used.}
\label{Appendix:table_mask_info}
\end{table}